\documentclass[conference]{IEEEtran}
\IEEEoverridecommandlockouts
\usepackage[style=ieee]{biblatex}
\usepackage{amsmath,amssymb,amsfonts}
\usepackage{algorithm,algorithmic}
\usepackage{graphicx}
\usepackage{textcomp}
\usepackage{multirow,multicol}
\usepackage{booktabs}
\usepackage{hyperref}

\bibliography{references.bib}

\begin{document}

\title{mil-benchmarks: Standardized Evaluation of Deep Multiple-Instance Learning Techniques}

\author{
    \IEEEauthorblockA{
    \IEEEauthorblockN{Daniel Grahn}
        \textit{Department of Computer Science and Engineering} \\
        \textit{Wright State University}\\
        Dayton, OH
        dan.grahn@wright.edu
    }
}

\maketitle

\begin{abstract}
Multiple-instance learning is a subset of weakly supervised learning where labels are applied to sets of instances rather than the instances themselves. Under the standard assumption, a set is positive only there is if at least one instance in the set which is positive.

This paper introduces a series of multiple-instance learning benchmarks generated from MNIST, Fashion-MNIST, and CIFAR10. These benchmarks test the standard, presence, absence, and complex assumptions and provide a framework for future benchmarks to be distributed. I implement and evaluate several multiple-instance learning techniques against the benchmarks. Further, I evaluate the Noisy-And method with label noise and find mixed results with different datasets. The models are implemented in TensorFlow 2.4.1 and are available on GitHub. The benchmarks are available from PyPi as \textit{mil-benchmarks} and on GitHub.
\end{abstract}

\begin{IEEEkeywords}
Multiple-instance learning, Weakly supervised learning, Artificial neural networks
\end{IEEEkeywords}

\section{Introduction}
Supervised learning is a large and well-studied branch of machine learning research. In the typical supervised learning dataset, each input instance has a corresponding label. Models are trained directly on the instances and labels. However, there are many applications where providing a label for each instance is not feasible due to data collection, scope, budget, or constraints. In order to handle these cases, supervised learning is weakened in different ways. One weakening of supervised learning is Multiple-Instance Learning (MIL). MIL does not provide labels for each input instance. Rather, instances are grouped into sets, commonly referred to as bags. Each of these bags is given a label based on its contents.

\citeauthor{babenko2008multiple} provides an useful example of MIL \cite{babenko2008multiple}. Consider a group of faculty who each have a key ring with several keys on it. You know which faculty are able to access a specific lab and which are not. The task is to build a classifier which can predict whether any given key, and therefore the key ring containing it, grants access to the lab.

\begin{figure}[ht]
\centerline{\includegraphics[width=\columnwidth]{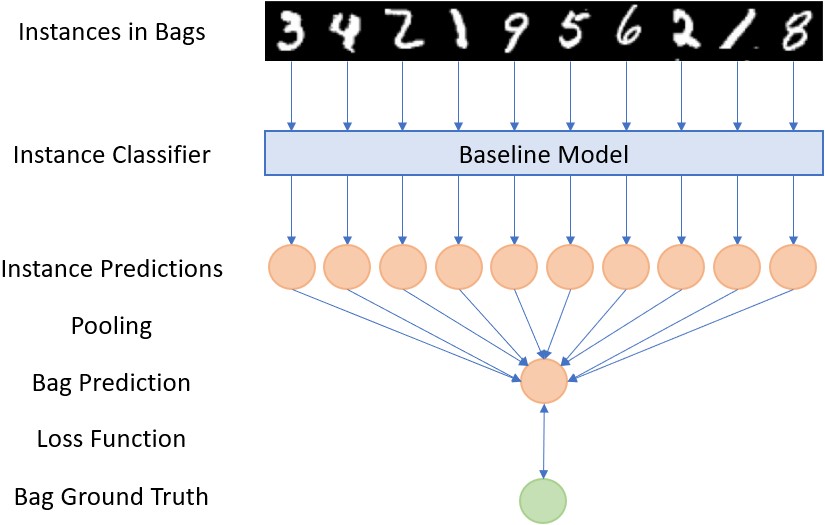}}
\caption{Example MIL Architecture}
\label{fig}
\end{figure}

This example uses the \textit{standard assumption} that a bag (key ring) is positive (grants access to the lab) if at least one instance (key) in the bag is positive. While not as common as the standard assumption, other assumptions are commonly used \cite{foulds2010review}. And while standard datasets exist for multiple instance learning \cite{musk1, musk2, ucsbcancer}, they largely use the standard assumption. In order to advance our understanding of MIL tasks, I propose a set of benchmarks which allow other assumptions to be tested on multiple datasets.

Additionally, many of the methods of multiple-instance learning are derived from traditional machine learning techniques \cite{carbonneau2018multiple}. As deep learning continues to grow in popularity and improve in performance, it will become more important to be able to use and evaluate deep-learning techniques on MIL tasks. To support this, I evaluate several MIL methods against the benchmarks.

This paper implements models in TensorFlow 2.4.1 \cite{abadi2016tensorflow}. Section \ref{datasets} describes the creation of the benchmark datasets. Section \ref{models} describes the MIL methods which have been implemented. The evaluation experiments and their results are presented and analyzed in section \ref{experiments}. And section \ref{conclusion} contains a summary and directions for future work.

\section{Benchmarks \label{datasets}}

Since traditional machine learning techniques work well with tabular data, the benchmarks are sourced from three computer vision datasets -- MNIST \cite{lecun1998gradient}, Fashion-MNIST \cite{xiao2017fashion}, and CIFAR-10 \cite{krizhevsky2009learning}.

I generate the datasets using Algorithm \ref{alg:dataset-gen}. An effect of this algorithm is that the datasets are not balanced. Table \ref{tab:ratios} contains a detailed breakdown of the ratio of positive instances in each benchmark dataset.

\subsection{Standard}
The standard assumption is that a bag will be labelled as positive if it contains at least one instance of a specific class of interest. The three selected datasets each have ten different classes. The classes have different difficulties for classification. To avoid choosing classes which are overly easy or hard as the class of interest, I generate datasets for each class. This provides 30 benchmarks with 10 from each source. If computationally feasible, evaluators should use all the benchmarks from a source and aggregate the results.

\subsection{Presence \& Absence}
The \textit{presence} assumption labels a bag as positive if any one of the members of a subset are present in the bag. For instance, a bag from MNIST may be positive if either a 0 or a 1 are present. The benchmarks include 9 presence assumptions from each dataset using classes of interest $(0, 1),(1,2)..(8,9)$.

The absence assumption is the negation of the presence. A bag is positive if it does \textit{not} include a subset of classes. The benchmarks include 9 absence benchmarks for the same classes of interest. Again, evaluators should prefer aggregate results across these benchmarks.

\subsection{Complex}
Two datasets are included with a \textit{complex} assumption and are based on whether an entire outfit is present in a bag. Table \ref{tab:outfit-mapping} contains the definitions of the outfits. The \textit{basic outfit} assumes that an outfit contains a top, bottom, and shoes but does not require a handbag. I consider coats as tops because the two classes are difficult to differentiate. I also included dresses with the tops because dresses may be worn with bottoms. The \textit{multi outfit} is stricter. It allows for two types of outfits: a t-shirt or shirt with trousers and sneakers or boots -- or -- a dress with a bag and sandals or boots.

\begin{table}[h]
\caption{Outfit Classes}
\begin{center}
\begin{tabular}{ll}
\toprule
\textbf{Outfit} & \textbf{Contents}                                                     \\
\midrule
Basic           & $(0 \vee 2 \vee 3 \vee 4 \vee 6) \wedge (1) \wedge (6 \vee 7 \vee 9)$ \\
\midrule
Multi           & $[(0 \vee 6) \wedge (1) \wedge (7 \vee 9)] \vee $                     \\
                & $[(3) \wedge (8) \wedge (5 \vee 9)]$                                  \\
\bottomrule
\end{tabular}
\end{center}
\label{tab:outfit-mapping}
\end{table}

\begin{algorithm}[ht]
\caption{Dataset Generation Algorithm}
\begin{algorithmic}[1]
    \renewcommand{\algorithmicrequire}{\textbf{Input:}}
    \renewcommand{\algorithmicensure}{\textbf{Output:}}
    \REQUIRE $x_{ix}$ ($x$ indices), $y$, $min\_bag=3$, $max\_bag=7$
    \ENSURE  $x_b$, $y_b$
    \STATE $x_s, y_s$ shuffle $x_{ix}$ and $y$
    \STATE $x_b, y_b = [], []$
    \STATE $i = 0$
    \WHILE {$|x_s| - i > max\_bag$}
        \STATE $s = $ random in range $[min\_bag, max\_bag]$
        \STATE $l = $ label $y[i..i+s]$ according to assumption
        \STATE append $x_{ix}[i..i+s]$ to $x_b$
        \STATE append $l$ to $y_b$
        \STATE $i = i + s$
    \ENDWHILE
    
    \STATE $l = $ label for $y[i..]$ according to assumption
    \STATE append $x_{ix}[i..]$ to $x_b$
    \STATE append $l$ to $y_b$

    \RETURN $x_b, y_b$
\end{algorithmic}
\label{alg:dataset-gen}.
\end{algorithm}

\begin{table*}[hbtp]
\caption{Dataset Ratios}
\begin{center}
\begin{tabular}{llllll}
\toprule
\textbf{Assumption} & \textbf{Positive If} & \textbf{Source} & \textbf{Num.} & \textbf{Avg. Train Ratio} & \textbf{Avg. Test Ratio} \\ \midrule
Standard & $x$ is present           & CIFAR-10 & $10$ & 40.11\% & 40.20\% \\
         &                          & Fashion  & $10$ & 40.25\% & 40.28\% \\
         &                          & MNIST    & $10$ & 40.23\% & 40.26\% \\
\midrule
Presence & $x$ or $x+1$ is present  & CIFAR-10 & $9$  & 65.33\% & 65.27\% \\
         &                          & Fashion  & $9$  & 65.45\% & 65.55\% \\
         &                          & MNIST    & $9$  & 65.56\% & 65.50\% \\
\midrule
Absence  & $x$ and $x+1$ are absent & CIFAR-10 & $9$  & 34.68\% & 34.78\% \\
         &                          & Fashion  & $9$  & 34.55\% & 34.45\% \\
         &                          & MNIST    & $9$  & 34.44\% & 34.50\% \\
\midrule
Complex  & Contains "Outfit"        & Fashion  & $2$  & 25.51\% & 25.50\% \\
\bottomrule
\end{tabular}
\end{center}
\label{tab:ratios}
\end{table*}

\section{Models \label{models}}
\begin{table*}[ht]
\caption{Baseline Model Architecture}
\begin{center}
\begin{tabular}{llll}
\toprule
\textbf{Layer} & \textbf{Output Shape} & \textbf{\# of Params} & \textbf{Details} \\ \midrule
Convolution    & (28, 28, 64)          & 320                   & filters=64, kernel\_size=2, padding='same', activation='relu'             \\
Max Pool       & (14, 14, 64)          & 0                     & pool\_size=2                                                              \\
Convolution    & (12, 12, 32)          & 18,464                & filters=32, kernel\_size=3, padding='valid', activation='relu'            \\
Max Pool       & (6, 6, 32)            & 0                     & pool\_size=2                                                              \\
Dropout        & (6, 6, 32)            & 0                     & dropout=0.3                                                               \\
Flatten        & (1152)                & 0                     &                                                                           \\
Dense          & (256)                 & 295,168               & activation='relu'                                                         \\
Dropout        & (256)                 & 0                     & dropout=0.5                                                               \\
Dense          & (10)                  & 2,570                 & activation='softmax'                                                      \\ \midrule
\multicolumn{4}{l}{
    \begin{tabular}[c]{@{}l@{}}
        Total Parameters: 316,522 \\
        Optimizer: Adam (default settings) \\
        Loss: Categorical Cross-Entropy \\
        Epochs: 10
    \end{tabular}
} \\
\bottomrule
\end{tabular}
\end{center}
\label{baseline-model}
\end{table*}
\begin{table*}[tbp]
\caption{Test Result Metrics}
\begin{center}
\begin{tabular}{llllllllllll}
\toprule

\textbf{Assumption} & \textbf{Source} & \multicolumn{2}{l}{\textbf{Fully-Connected}} & \multicolumn{2}{l}{\textbf{Max Pool}} & \multicolumn{2}{l}{\textbf{Max Pool + FC}} & \multicolumn{2}{l}{\textbf{Avg Pool + FC}} & \multicolumn{2}{l}{\textbf{Noisy-And}} \\
 & &  \textit{F1} & \textit{AUC} & \textit{F1} & \textit{AUC} & \textit{F1} & \textit{AUC} & \textit{F1} & \textit{AUC} & \textit{F1} & \textit{AUC} \\
\midrule
Standard & CIFAR-10     & 0.533 & 0.679 & 0.546          & 0.701 & 0.492 & 0.663 & 0.476 & 0.662 & \textbf{0.766} & \textbf{0.851} \\
         & Fashion      & 0.848 & 0.849 & 0.632          & 0.770 & 0.728 & 0.820 & 0.550 & 0.724 & \textbf{0.935} & \textbf{0.972} \\
         & MNIST        & 0.803 & 0.881 & 0.742          & 0.795 & 0.669 & 0.769 & 0.660 & 0.797 & \textbf{0.991} & \textbf{0.997} \\
\midrule
Presence & CIFAR-10     & 0.539 & 0.712 & 0.463          & 0.710 & 0.494 & 0.716 & 0.519 & 0.715 & \textbf{0.624} & \textbf{0.776} \\
         & Fashion      & 0.701 & 0.834 & \textbf{0.900} & 0.829 & 0.504 & 0.734 & 0.635 & 0.795 & 0.894          & \textbf{0.954} \\
         & MNIST        & 0.824 & 0.905 & 0.975          & 0.873 & 0.566 & 0.768 & 0.546 & 0.760 & \textbf{0.981} & \textbf{0.995} \\
\midrule
Absence  & CIFAR-10     & 0.572 & 0.732 & 0.462          & 0.712 & 0.486 & 0.707 & 0.491 & 0.714 & \textbf{0.622} & \textbf{0.774} \\
         & Fashion      & 0.675 & 0.810 & 0.899          & 0.830 & 0.518 & 0.738 & 0.667 & 0.818 & \textbf{0.889} & \textbf{0.952} \\
         & MNIST        & 0.867 & 0.930 & 0.971          & 0.855 & 0.658 & 0.818 & 0.586 & 0.774 & \textbf{0.980} & \textbf{0.996} \\
\midrule
Complex  & Basic Outfit & 0.416 & 0.713 & 0.798          & 0.764 & 0.416 & 0.713 & 0.416 & 0.713 & \textbf{0.867} & \textbf{0.948} \\
Complex  & Multi Outfit & 0.849 & 0.944 & 0.764          & 0.799 & 0.416 & 0.713 & 0.416 & 0.713 & \textbf{0.867} & \textbf{0.945} \\
\bottomrule
\end{tabular}
\end{center}
\label{metrics}
\end{table*}

\subsection{Baseline Model}
Before tackling the multiple-instance learning techniques, I train a baseline model which can separate MNIST, Fashion-MNIST, and CIFAR-10. Perfect performance is not necessary. Given my limited computational resources, it is more important to have a lightweight model than state-of-the-art performance.

To build this baseline model, I perform minimal modification of the input data. The input images are normalized to $[0..1]$ rather than $[0..255]$. And the classes are converted from nominal to one-hot representation. Table \ref{baseline-model} shows the entire configuration of the baseline model. The model trained on CIFAR-10 receives an F1 of $0.718$ and an AUC of $0.963$. On Fashion-MNIST it reports an F1 of $0.921$ and an AUC of $0.9961$. And on MNIST, an F1 of $0.992$ and an AUC of $0.999$. When run on CIFAR-10, the input shape is changed but all other layers are left identical.

\subsection{Adapting for MIL}
Before running the baseline model with MIL methods, a few modifications must be made. The simplest modification is the number of output classes. Instead of having 10 output classes, the MIL model has 2. This is changed in the baseline model architecture.

The baseline model is designed to accept one image at a time and produce a single layer. The MIL models must accept $3..7$ different images and produce a single prediction. To accomplish this, each bag is padded with zeros so that it has the same size as a bag with seven instances. While this is not strictly necessary, it allows training in batches rather than on a single bag at a time.

Next, each layer from the baseline model is wrapped in \texttt{tensorflow.keras.layers.TimeDistributed}. This allows each instance within the bags to propagate through the baseline network with identical weights. It minimizes the number of parameters to be trained and also enforces that the position of the instances in the bags is irrelevant. 

With the above described modifications, the model makes a prediction for each instance in the bag. The different MIL methods used are focused on taking instance-level predictions and merging them into a bag-level prediction.

\subsection{MIL Models}
I selected five MIL models to evaluate. They are:

\begin{enumerate}
    \item Fully-Connected Layer\label{dense}
    \item Max Pool \cite{su2017weakly,kumar2016audio,salamon2017dcase} \label{max}
    \item Max Pool + Fully-Connected Layer \label{max+dense}
    \item Avg Pool + Fully-Connected Layer \label{avg+dense}
    \item Noisy-And \cite{kraus2016classifying} \label{noisy-and}
\end{enumerate}

For Model \ref{dense} (Fully Connected), the instance predictions are flattened and fully connected to the outputs. Model \ref{max} (Max Pool) simply applies a 1D max pooling to the output predictions. No additional weights are applied. Model \ref{max+dense} is an extension of Model \ref{max} where a fully-connected layer is added after the max pool. Model \ref{avg+dense} modifies Model \ref{max+dense} by replacing the Max Pooling with an average pooling. Finally, Model \ref{noisy-and} uses a custom layer to apply the noisy-And function defined in Equation \ref{eq:noisy-and}. 

\begin{align}
    && P_i &= g_i(\{p_{ij}\}) = \frac{\sigma(a(p_{\overline{ij}} - b_i)) - \sigma(-ab_i)}{\sigma(a(1 - b_i)) - \sigma(-ab_i)} \label{eq:noisy-and} \\
    \text{Where} && p_{\overline{ij}} &= \frac{1}{|j|} \sum_j p_{ij}
\end{align}

The Noisy-And function imitates a probabilistic And. It triggers a bag-level positive prediction ($P_i$) when the mean of instance-level probabilities ($P_{\overline{ij}}$) crosses a learned threshold. $a$ controls the slope of the activation function. The authors find the best performance at $a=10$ which we adopt. $b_i$ are parameters which adjust the threshold for each class.\footnote{Under some training circumstances, predictions for this layer fall outside the $[0..1]$ prediction range. The predictions are clipped to avoid this.}

To keep the training as consistent as possible, random seeds are always set to $42$ \cite{adams2017ultimate} and no modifications are made to loss functions, optimizers, data preprocessing, or any other hyperparameters.

\section{Experiments \label{experiments}}

\begin{figure*}[t]
\centerline{\includegraphics[width=6in]{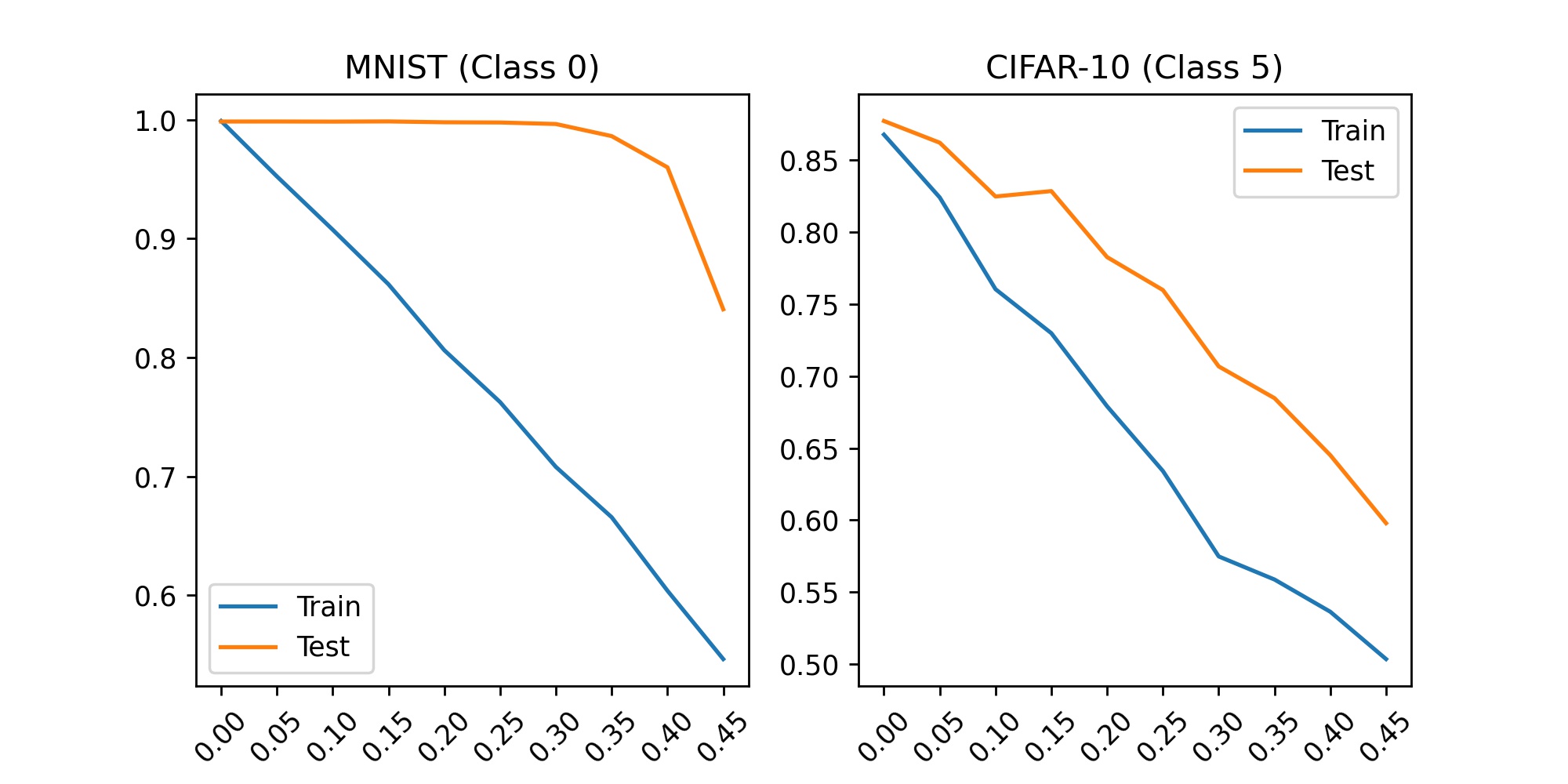}}
\caption{Noisy-And Performance on Noisy Labels}
\label{fig:noisy-labels}
\end{figure*}

\subsection{MIL Task}
Table \ref{metrics} contains metrics for each model trained on each dataset. For many of the benchmarks, Model \ref{dense} fails to learn. It ends up predicting negative for all bags. This failure to converge even happens within datasets and assumptions. For instance, the model learns the standard assumption for MNIST when the class of interest is $[2,3,4,6,7,8,9]$ but not for $[0,1,5]$. The model learned the multi-outfit assumption but failed on the basic outfit. This reflects an instability in the method. It fails to learn the \textit{Basic Outfit} benchmark at all.

Model \ref{max} (Max Pool) was able to learn from the benchmarks. However, it was not always able to generalize well. Several of the benchmarks posted high training scores but low test scores. Additionally, there was an epoch to epoch stochasticity in the F1 and AUC metrics rather than a (more or less) monotonically increasing value. Despite posting lower scores than Model \ref{dense}, Model \ref{max} learns more consistently.

When a dense layer is added after the pooling layer (Model \ref{max+dense}), the downsides of both Model \ref{dense} and Model \ref{max} are diminished, but are not entirely mitigated. Model \ref{max+dense} fails to converge on more datasets than Model \ref{max}, but does have significantly less of a gap between training and test metrics in most cases. However, the model fails to learn from the Complex benchmark and posts the minimum scores of $0.416$ F1 and $0.713$ AUC.

Swapping the max pool layer for an avg pool (Model \ref{avg+dense}) has mixed results. Sometimes it performs better, sometimes it performs worse. In most cases it has no effect. On some MIL tasks it may be useful to swap Max for Avg pooling layers in an attempt to improve performance. Again, the model failed to learn from the Complex benchmark.

Finally, Model \ref{noisy-and} (Noisy-And) is extremely successful. It learns well on all of the datasets and posts the highest results on nearly all benchmarks. The layer has been designed specifically to solve MIL tasks and its results show that well.

\subsection{MIL with Noise}
Model \ref{noisy-and} performs well on all of the datasets. In order to better understand the capabilities of the Noisy-And network, I adapt the benchmarks to include noise in the labels. This is accomplished by randomly flipping classes with a probability equal to the level of noise. I then retrain the model on noise levels $[0, 0.45]$ incremented by $0.05$ with a single benchmark from the MNIST and CIFAR-10 sources (classes 0 and 5 respectively) and assess the performance at each step. AUC is an appropriate metric for this evaluation as it is insensitive to changes in class distribution \cite{flach2015precision}. The results are displayed in Figure \ref{fig:noisy-labels}. 

For the MNIST dataset, the model learns with only slight degradation of performance until the noise level passes $0.35$. Even with $~45\%$ of the labels flipped, the model receives an AUC of $0.706$. Training on the CIFAR-10 dataset produces different results. The AUC degrades roughly in line with the amount of noise which has been added to the labels. I hypothesize that the baseline network with the Noisy-And layer is extremely capable of classifying MNIST digits and thus can handle large amounts of noise. On the other hand, the model scores lower on CIFAR-10 without label noise and because of this we see the drop in AUC right away. More testing is necessary to confirm his hypothesis.

\section{Conclusion \label{conclusion}}
This paper makes three contributions. First, it creates and distributes 86 MIL benchmark datasets. These datasets span three sources and provide a means to benchmark assumptions which are not covered by existing datasets. In addition, I have released a Python module which provides easy access to these datasets and is available on PyPi as \textit{mil-benchmarks} at \href{https://pypi.org/project/mil-benchmarks/}{https://pypi.org/project/mil-benchmarks/}. This module provides a platform for distribution of future MIL benchmarks.

Currently, the \textit{mil-benchmarks} module stores datasets as CSVs. Future work may reduce the storage size of the module by recreating these CSVs using reproducible random numbers. Additional benchmarks may be created from more complicated source data, such as:

\begin{enumerate}
    \item CIFAR-100 \cite{krizhevsky2009learning}
    \item Street View House Numbers \cite{netzer2011reading}
    \item miniImageNet \cite{vinyals2016matching}
\end{enumerate}

Second, five methods were evaluated against the MIL benchmarks. These evaluations show the inadequacy of standard fully-connected and standard pooling layers for use in MIL task domain. But, they show that the Noisy-And MIL layer performs much better while using the same underlying classification model.

Future research may explore more assumptions under which bags may be labelled. Assumptions of interest include labeling bags as positive if:

\begin{enumerate}
    \item A class's representation crosses some threshold.
    \item One or another class is present, but not both.
    \item More complicated boolean expressions.
\end{enumerate}

Finally, Noisy-And was found to produce mixed results under different circumstances. I hypothesize that this is related to the performance of the model on a noise-free dataset
be highly-effective in the presence of noisy labels.

The models, layer implementations, and Jupyter Notebooks to reproduce the results of this paper are available on GitHub at \href{https://github.com/dgrahn/deep-mil}{https://github.com/dgrahn/deep-mil}.

\printbibliography

\end{document}